\documentclass[10pt,twocolumn,letterpaper]{article}

\usepackage{cvpr}
\usepackage{times}
\usepackage{epsfig}
\usepackage{graphicx}
\usepackage{amsmath}
\usepackage{amssymb}
\usepackage{enumerate}
\usepackage{makecell}

\usepackage[noend]{algpseudocode}
\usepackage[]{algorithm2e}
\usepackage{caption}
\usepackage{subcaption}
\usepackage{float}
\usepackage{dblfloatfix}
\usepackage[font={small}]{caption}
\usepackage{amssymb}
\usepackage{pifont}
\newcommand{\cmark}{\ding{51}}
\newcommand{\xmark}{\ding{55}}
\usepackage{pdfpages}
\usepackage{multirow}
\usepackage{url}

\usepackage[pagebackref=true,breaklinks=true,letterpaper=true,colorlinks,bookmarks=false]{hyperref}

\cvprfinalcopy 

\def\cvprPaperID{4972} 

\ifcvprfinal\fi
\begin{document}

\title{SpineNet: Learning Scale-Permuted Backbone for Recognition and Localization}

\author{
Xianzhi Du\qquad
Tsung-Yi Lin\qquad
Pengchong Jin\qquad
Golnaz Ghiasi\\
Mingxing Tan\qquad
Yin Cui\qquad
Quoc V. Le\qquad
Xiaodan Song\\
Google Research, Brain Team\\
{\tt\small \{xianzhi,tsungyi,pengchong,golnazg,tanmingxing,yincui,qvl,xiaodansong\}@google.com}
}

\maketitle
\thispagestyle{empty}

\begin{abstract}
Convolutional neural networks typically encode an input image into a series of intermediate features with decreasing resolutions. While this structure is suited to classification tasks, it does not perform well for tasks requiring simultaneous recognition and localization (\eg, object detection). The encoder-decoder architectures are proposed to resolve this by applying a decoder network onto a backbone model designed for classification tasks. In this paper, we argue encoder-decoder architecture is ineffective in generating strong multi-scale features because of the scale-decreased backbone. We propose SpineNet, a backbone with scale-permuted intermediate features and cross-scale connections that is learned on an object detection task by Neural Architecture Search. 
Using similar building blocks, SpineNet models outperform ResNet-FPN models by $\sim$3\% AP at various scales while using 10-20\% fewer FLOPs. In particular, SpineNet-190 achieves \textbf{52.5\% AP} with a Mask R-CNN detector and achieves \textbf{52.1\% AP} with a RetinaNet detector on COCO for a single model without test-time augmentation, significantly outperforms prior art of detectors. SpineNet can transfer to classification tasks, achieving 5\% top-1 accuracy improvement on a challenging iNaturalist fine-grained dataset. Code is at: \url{https://github.com/tensorflow/tpu/tree/master/models/official/detection}.
\end{abstract}

\begin{figure}
    \includegraphics[width=1.0\columnwidth]{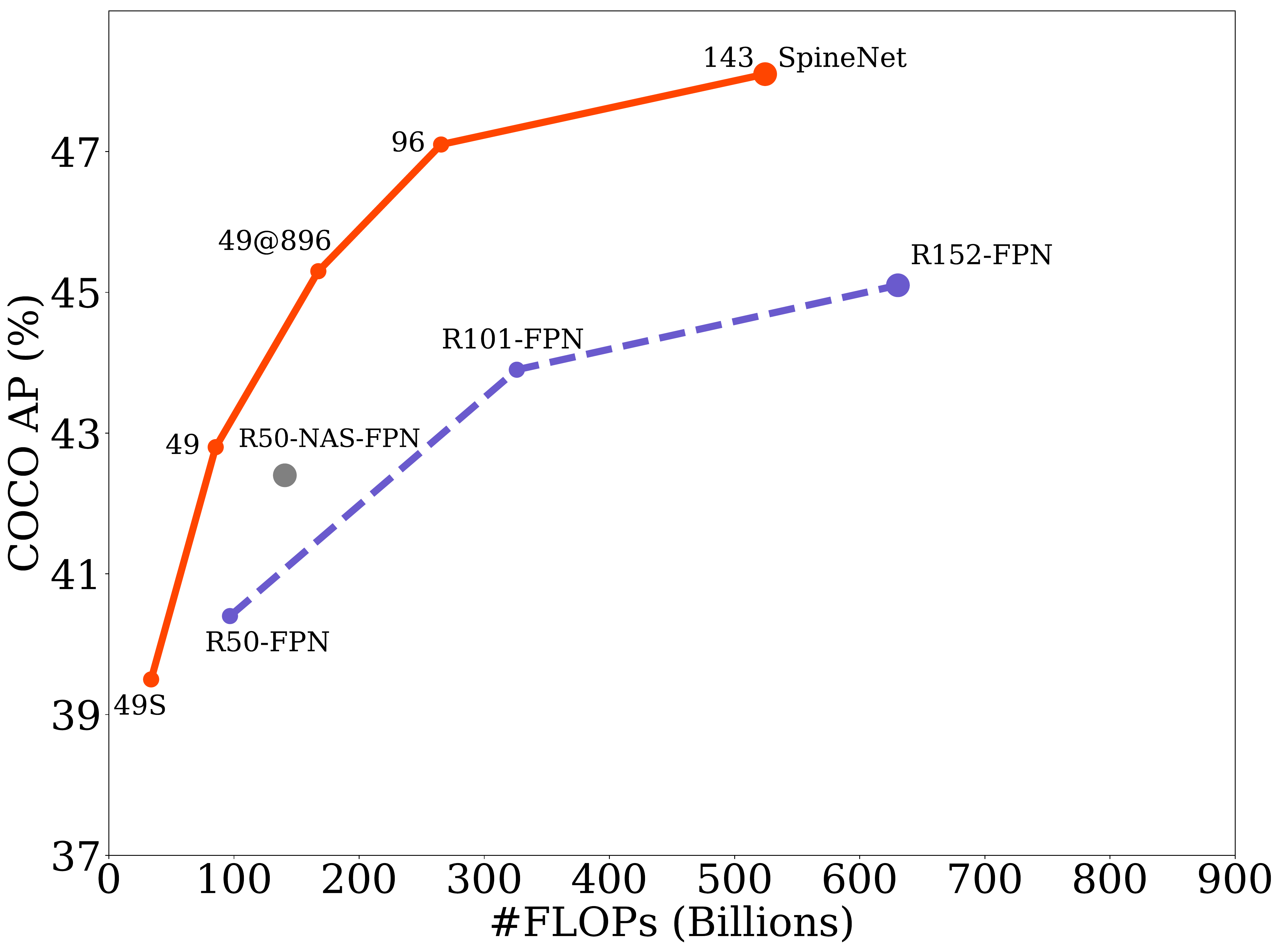}
    \hspace{-43mm}\resizebox{.46\columnwidth}{!}{
    \begin{tabular}[b]{l | c | c }
     & \multicolumn{1}{l|}{\#FLOPs$\blacktriangle$} & \multicolumn{1}{c}{AP}\\
    \Xhline{1.0pt}
    SpineNet-49S & 33.8B  & 39.5\\
    \hline
    SpineNet-49 & 85.4B &  42.8\\
    R50-FPN & 96.8B & 40.4 \\
    R50-NAS-FPN & 140.0B & 42.4 \\
    \hline
    SpineNet-49 @896 &  167.4B & 45.3\\
    \hline
    SpineNet-96 &  265.4B &  47.1\\
    R101-FPN & 325.9B & 43.9 \\
    \hline
    SpineNet-143 &  524.4B &  48.1\\
    R152-FPN & 630.5B & 45.1
    \vspace{11mm}
    \end{tabular}}
    \caption{The comparison of RetinaNet models adopting SpineNet, ResNet-FPN, and NAS-FPN backbones. Details of training setup is described in Section~\ref{sec:exp} and controlled experiments can be found in Table~\ref{tab:mainresults}, \ref{tab:evolveresults}.}
    \label{fig:detection_performance}
\vspace*{-2mm}
\end{figure}
\begin{figure}
    \includegraphics[width=1.0\columnwidth]{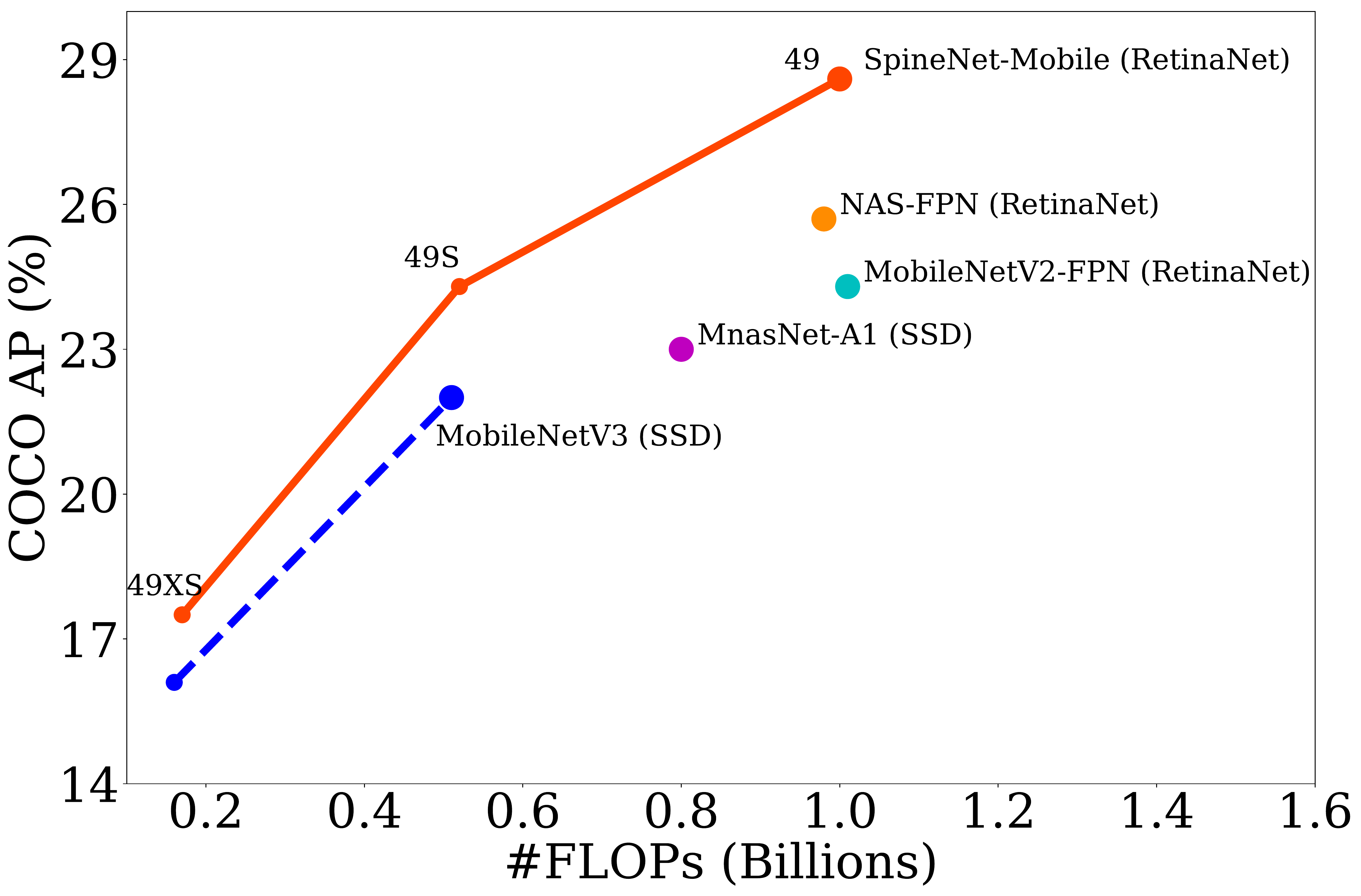}
    \hspace{-44mm}\resizebox{.47\columnwidth}{!}{
    \begin{tabular}[b]{l | r | l }
     & \multicolumn{1}{c|}{\#FLOPs$\blacktriangle$} & \multicolumn{1}{c}{AP} \\
    \Xhline{1.0pt}
    SpineNet-49XS & 0.17B  & 17.5 \\
    SpineNet-49S & 0.52B &  24.3\\
    SpineNet-49 &  1.00B & 28.6
    \vspace{11mm}
    \end{tabular}}
    \caption{A comparison of mobile-size SpineNet models and other prior art of detectors for mobile-size object detection. Details are in Table~\ref{tab:ondeviceresults}.}
    \label{fig:mobile_detection_performance}
\vspace*{-5mm}
\end{figure}

\vspace*{-0mm}
\section{Introduction}
In the past a few years, we have witnessed a remarkable progress in deep convolutional neural network design. Despite networks getting more powerful by increasing depth and width~\cite{resnet,zagoruykoK16wideresnet}, the meta-architecture design has not been changed since the invention of convolutional neural networks. Most networks follow the design that encodes input image into intermediate features with monotonically decreased resolutions. Most improvements of network architecture design are in adding network depth and connections within feature resolution groups~\cite{lecun1989backpropagation,resnet,densenet,nasnet}. LeCun~\etal~\cite{lecun1989backpropagation} explains the motivation behind this scale-decreased architecture design: \textit{``High resolution may be needed to detect the presence of a feature, while its exact position need not to be determined with equally high precision.''}

The scale-decreased model, however, may not be able to deliver strong features for multi-scale visual recognition tasks where recognition and localization are both important (\eg, object detection and segmentation). Lin et al.~\cite{fpn} shows directly using the top-level features from a scale-decreased model does not perform well on detecting small objects due to the low feature resolution. Several work including~\cite{fpn,chen2018encoderdecoder} proposes multi-scale encoder-decoder architectures to address this issue.
A scale-decreased network is taken as the encoder, which is commonly referred to a \textit{backbone} model.
Then a decoder network is applied to the backbone to recover the feature resolutions. The design of decoder network is drastically different from backbone model.
A typical decoder network consists of a series of cross-scales connections that combine low-level and high-level features from a backbone to generate strong multi-scale feature maps.
Typically, a backbone model has more parameters and computation (\eg, ResNets~\cite{resnet}) than a decoder model (\eg, feature pyramid networks~\cite{fpn}). Increasing the size of backbone model while keeping the decoder the same is a common strategy to obtain stronger encoder-decoder model.

\begin{figure}[t]
\centering
\includegraphics[width=0.9\linewidth]{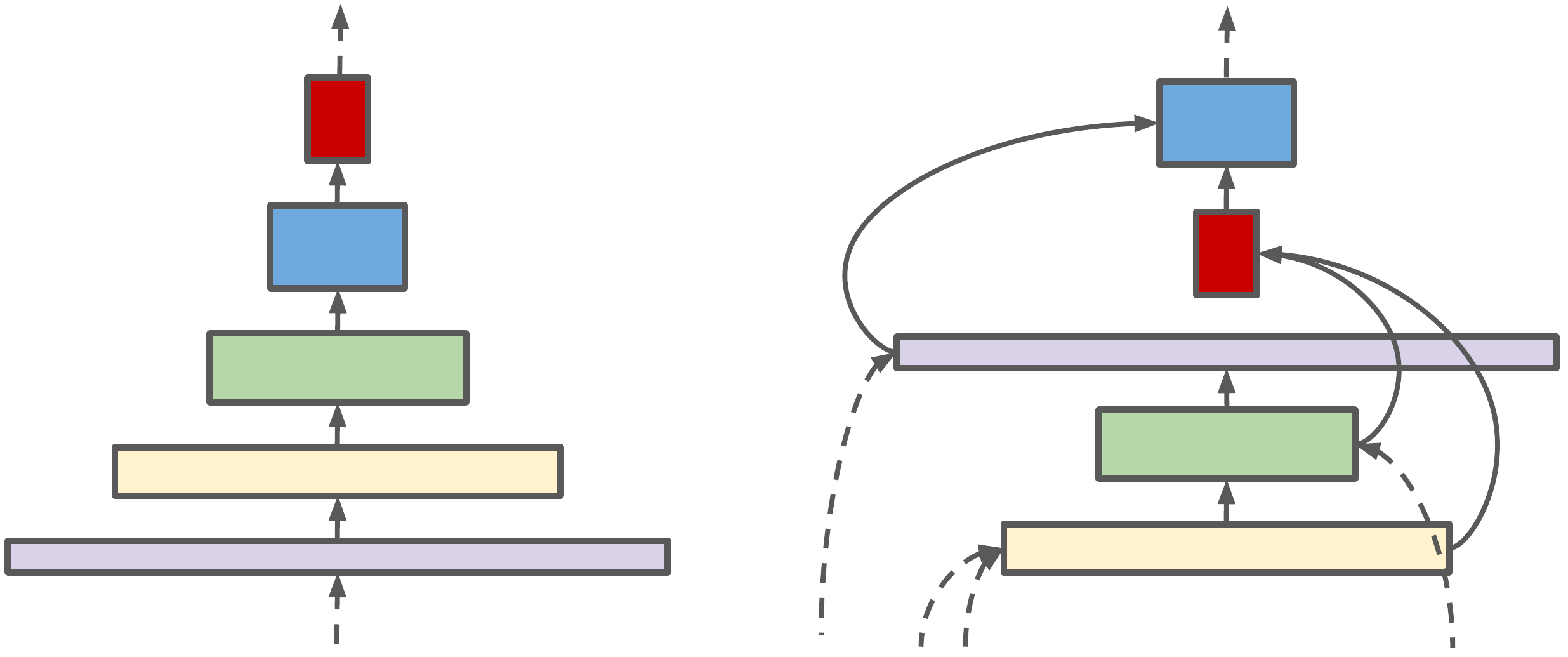}
\caption{An example of scale-decreased network (left) \vs scale-permuted network (right). The width of block indicates feature resolution and the height indicates feature dimension. Dotted arrows represent connections from/to blocks not plotted.}
\label{fig:toy_architecture}
\end{figure}

In this paper, we aim to answer the question: Is the scale-decreased model a good backbone architecture design for simultaneous recognition and localization? Intuitively, a scale-decreased backbone throws away the spatial information by down-sampling, making it challenging to recover by a decoder network. In light of this, we propose a meta-architecture, called scale-permuted model, with two major improvements on backbone architecture design. First, the scales of intermediate feature maps should be able to increase or decrease anytime so that the model can retain spatial information as it grows deeper. Second, the connections between feature maps should be able to go across feature scales to facilitate multi-scale feature fusion. Figure~\ref{fig:toy_architecture} demonstrates the differences between scale-decreased and scale-permuted networks.

Although we have a simple meta-architecture design in mind, the possible instantiations grow combinatorially with the model depth. To avoid manually sifting through the tremendous amounts of design choices, we leverage Neural Architecture Search (NAS)~\cite{nas} to learn the architecture. 
The backbone model is learned on the object detection task in COCO dataset~\cite{coco}, which requires simultaneous recognition and localization. Inspired by the recent success of NAS-FPN~\cite{nasfpn}, we use the simple one-stage RetinaNet detector~\cite{retinanet} in our experiments. In contrast to learning feature pyramid networks in NAS-FPN, we learn the backbone model architecture and directly connect it to the following classification and bounding box regression subnets. In other words, we remove the distinction between backbone and decoder models. The whole backbone model can be viewed and used as a feature pyramid network.

Taking ResNet-50~\cite{resnet} backbone as our baseline, we use the bottleneck blocks in ResNet-50 as the candidate feature blocks in our search space. We learn (1) the permutations of feature blocks and (2) the two input connections for each feature block. All candidate models in the search space have roughly the same computation as ResNet-50 since we just permute the ordering of feature blocks to obtain candidate models. The learned scale-permuted model outperforms ResNet-50-FPN by (\textit{+2.9\% AP}) in the object detection task. The efficiency can be further improved (\textit{-10\% FLOPs}) by adding search options to adjust scale and type (\eg, residual block or bottleneck block) of each candidate feature block. We name the learned scale-permuted backbone architecture SpineNet. Extensive experiments demonstrate that scale permutation and cross-scale connections are critical for building a strong backbone model for object detection. Figure~\ref{fig:detection_performance} shows comprehensive comparisons of SpineNet to recent work in object detection.

We further evaluate SpineNet on ImageNet and iNaturalist classification datasets. Even though SpineNet architecture is learned with object detection, it transfers well to classification tasks. Particularly, SpineNet outperforms ResNet by 5\% top-1 accuracy on iNaturalist fine-grained classification dataset, where the classes need to be distinguished with subtle visual differences and localized features. The ability of directly applying SpineNet to classification tasks shows that the scale-permuted backbone is versatile and has the potential to become a unified model architecture for many visual recognition tasks.

\begin{figure*}[t]
\hspace{-2mm}
\begin{subfigure}{.245\textwidth}
  \centering
  \captionsetup{justification=centering}
  \includegraphics[width=1.5\linewidth,angle=90,origin=c]{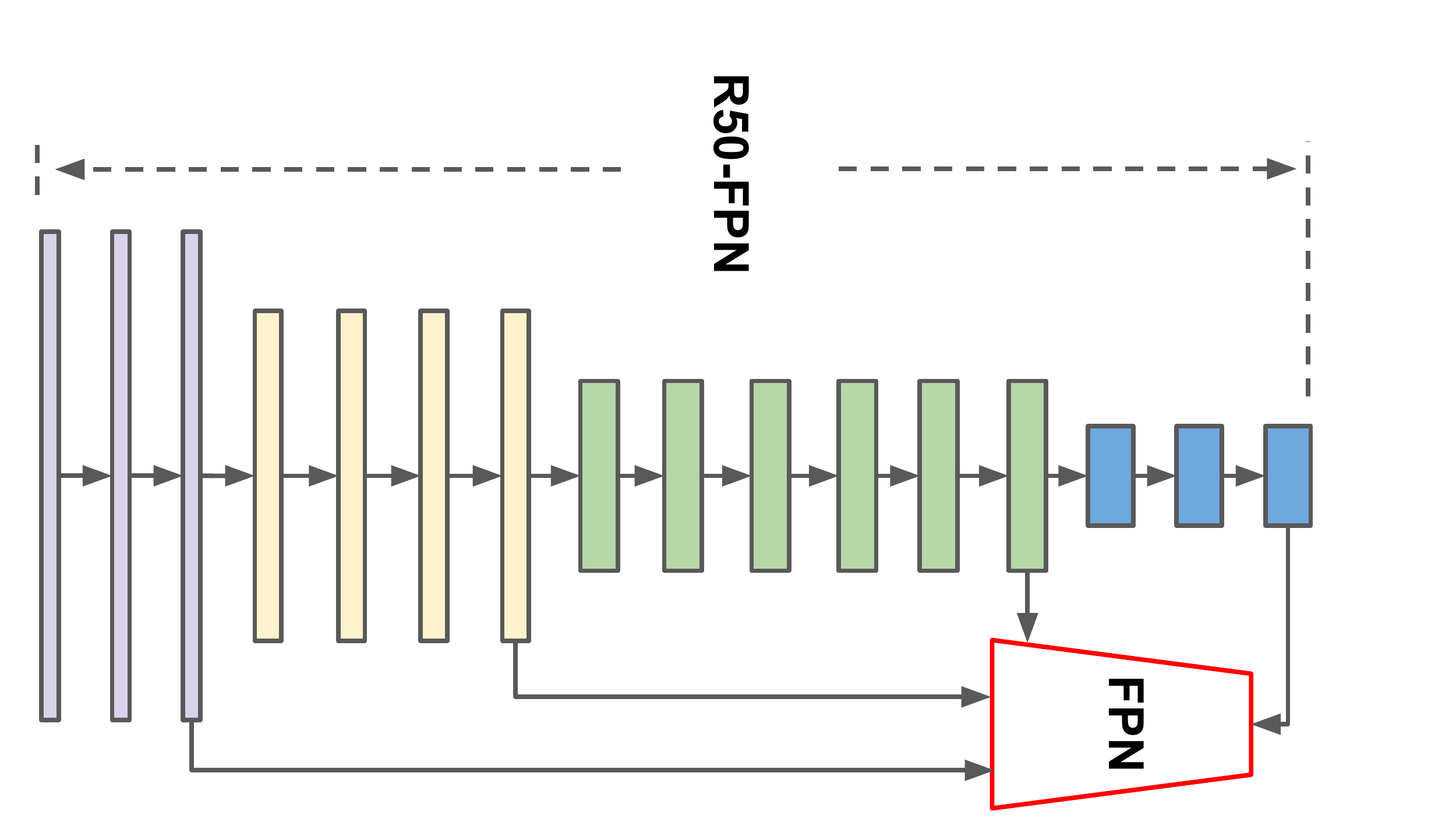} 
  \caption{R50-FPN @37.8\% AP}
  \label{fig:sub-first}
\end{subfigure}
\begin{subfigure}{.245\textwidth}
  \centering
  \captionsetup{justification=centering}
  \includegraphics[width=1.5\linewidth,angle=90,origin=c]{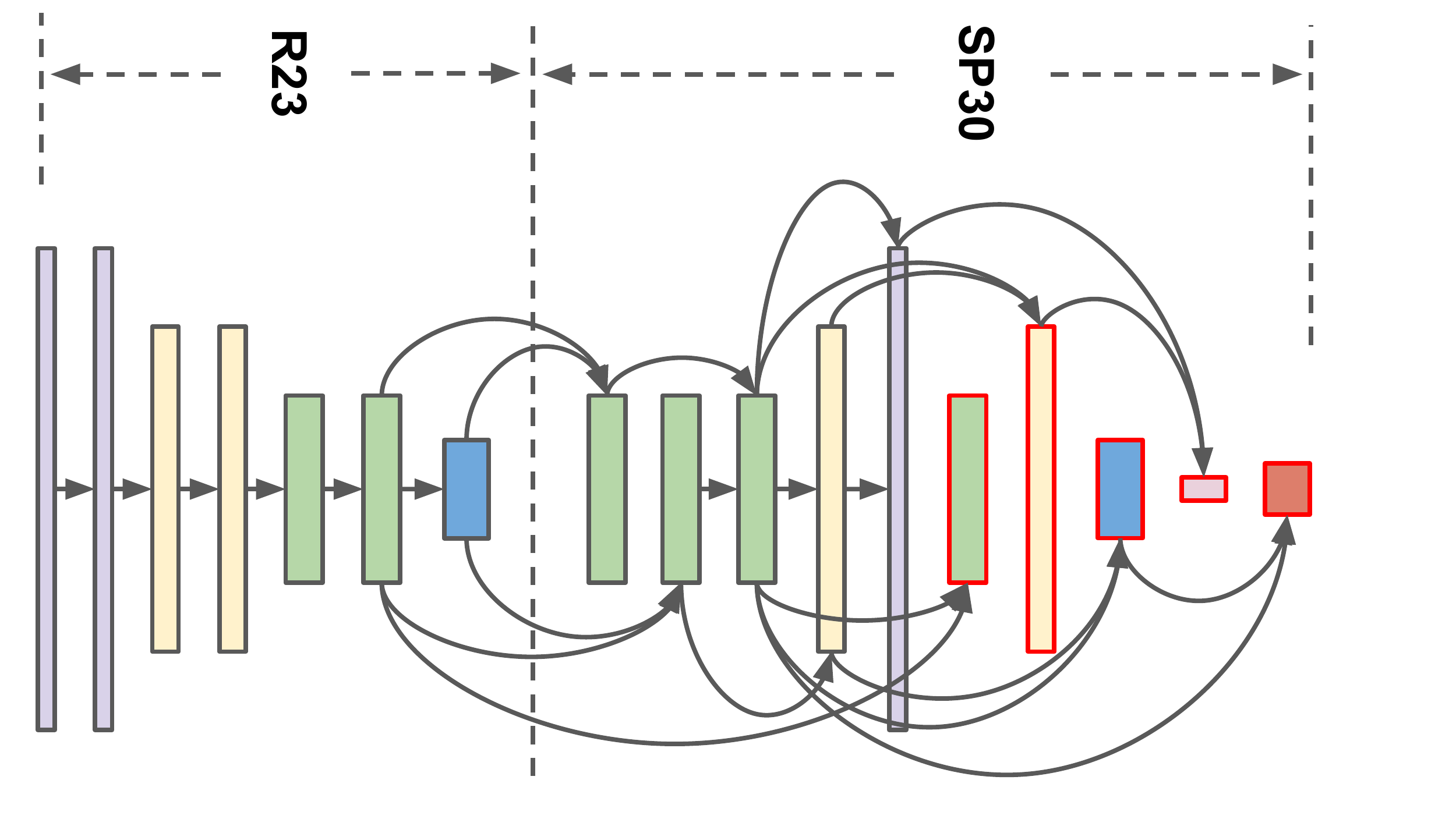} 
  \caption{R23-SP30 @39.6\% AP}
  \label{fig:sub-second}
\end{subfigure}
\begin{subfigure}{.245\textwidth}
  \centering
  \captionsetup{justification=centering}
  \includegraphics[width=1.5\linewidth,angle=90,origin=c]{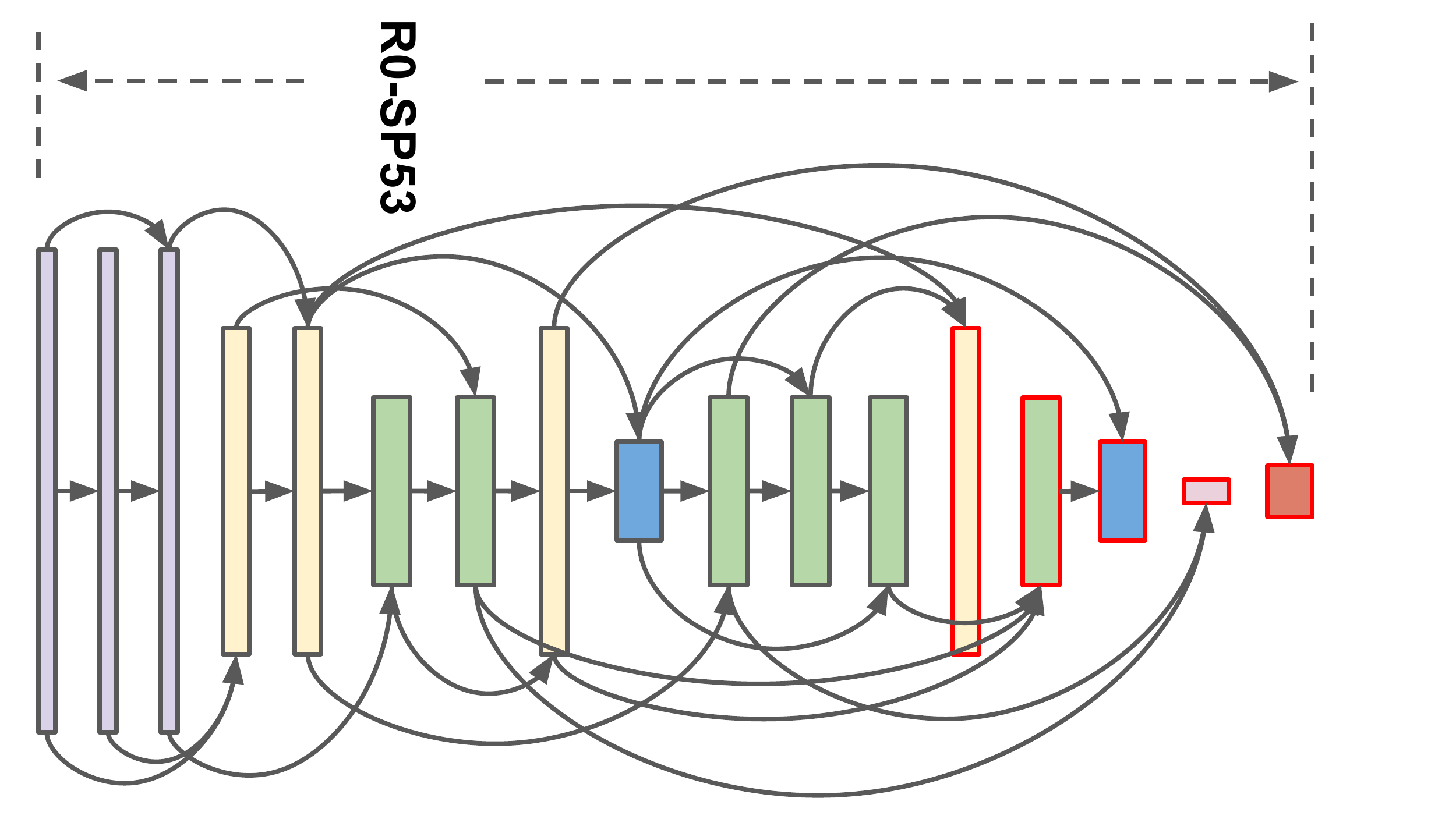} 
  \caption{R0-SP53 @40.7\% AP}
  \label{fig:sub-third}
\end{subfigure}
\begin{subfigure}{.245\textwidth}
  \centering
  \captionsetup{justification=centering}
  \includegraphics[width=1.5\linewidth,angle=90,origin=c]{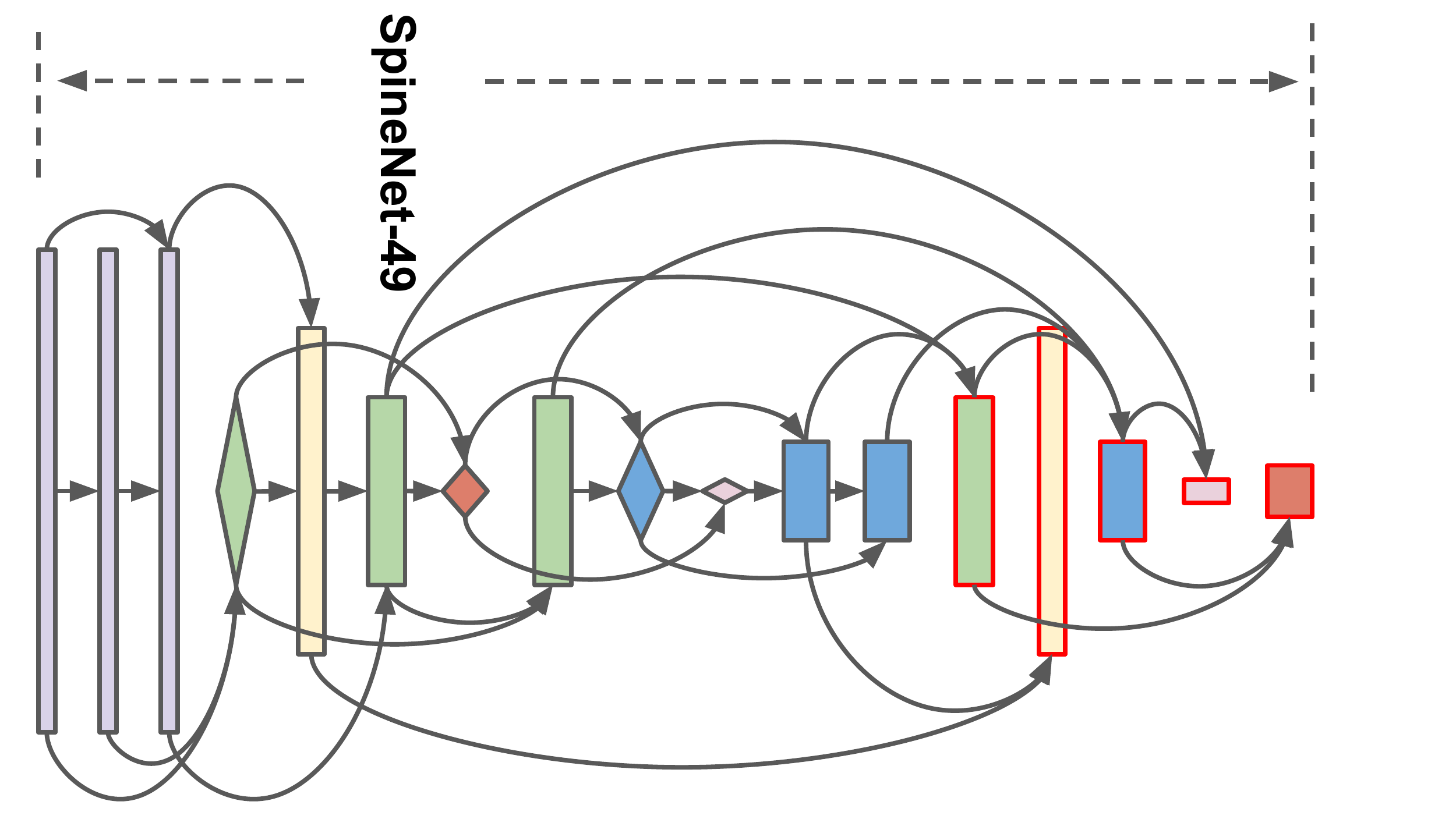}  
  \caption{SpineNet-49 @40.8\% AP}
  \label{fig:sub-forth}
\end{subfigure}

\caption{\textbf{Building scale-permuted network by permuting ResNet.} 
From (a) to (d), the computation gradually shifts from ResNet-FPN to scale-permuted networks. (a) The R50-FPN model, spending most computation in ResNet-50 followed by a FPN, achieves 37.8\% AP; (b) R23-SP30, investing 7 blocks in a ResNet and 10 blocks in a scale-permuted network, achieves 39.6\% AP; (c) R0-SP53, investing all blocks in a scale-permuted network, achieves 40.7\% AP; (d) The SpineNet-49 architecture achieves 40.8\% AP with 10\% fewer FLOPs (85.4B \vs 95.2B) by learning additional block adjustments. Rectangle block represent bottleneck block and diamond block represent residual block. Output blocks are indicated by red border.}
\label{fig:architectures}
\vspace{-2mm}
\end{figure*}

\section{Related Work}
\subsection{Backbone Model}
The progress of developing convolutional neural networks has mainly been demonstrated on ImageNet classification dataset~\cite{deng09imagenet}. Researchers have been improving model by increasing network depth~\cite{alexnet}, novel network connections~\cite{resnet,inceptionv1,inceptionv3,inceptionv4,densenet,senet}, enhancing model capacity~\cite{zagoruykoK16wideresnet,gpipe} and efficiency~\cite{xception,mobilenetv2,mobilenetv3,tan2019efficientnet}. Several works have demonstrated that using a model with higher ImageNet accuracy as the backbone model achieves higher accuracy in other visual prediction tasks~\cite{huang2017speedaccuracy,fpn,chen2018encoderdecoder}.

However, the backbones developed for ImageNet may not be effective for localization tasks, even combined with a decoder network such as~\cite{fpn,chen2018encoderdecoder}. DetNet~\cite{detnet} argues that down-sampling features compromises its localization capability. HRNet~\cite{hrnet} attempts to address the problem by adding parallel multi-scale inter-connected branches. Stacked Hourglass~\cite{stackedhourglass} and FishNet~\cite{fishnet} propose recurrent down-sample and up-sample architecture with skip connections. Unlike backbones developed for ImageNet, which are mostly scale-decreased, several works above have considered backbones built with both down-sample and up-sample operations. In Section~\ref{sec:ablation} we compare the scale-permuted model with Hourglass and Fish shape architectures.

\subsection{Neural Architecture Search}
Neural Architecture Search (NAS) has shown improvements over handcrafted models on image classification in the past few years~\cite{nasnet, pnasnet, darts, randomw, amoebanet, tan2019efficientnet}. Unlike handcrafted networks, NAS learns architectures in the given search space by optimizing the specified rewards. Recent work has applied NAS for vision tasks beyond classification. NAS-FPN~\cite{nasfpn} and Auto-FPN~\cite{xu2019autofpn} are pioneering works to apply NAS for object detection and focus on learning multi-layer feature pyramid networks. DetNAS~\cite{detnas} learns the backbone model and combines it with standard FPN~\cite{fpn}. Besides object detection, Auto-DeepLab~\cite{autodeeplab} learns the backbone model and combines it with decoder in DeepLabV3~\cite{chen2018encoderdecoder} for semantic segmentation. 
All aforementioned works except Auto-DeepLab learn or use a scale-decreased backbone model for visual recognition.

\section{Method}

The architecture of the proposed backbone model consists of a fixed stem network followed by a learned scale-permuted network. A stem network is designed with scale-decreased architecture. Blocks in the stem network can be candidate inputs for the following scale-permuted network.

A scale-permuted network is built with a list of building blocks $\{\mathbf{B}_1, \mathbf{B}_2,\cdots, \mathbf{B}_N\}$. Each block $\mathbf{B}_k$ has an associated feature level $L_i$. Feature maps in an $L_i$ block have a resolution of $\frac{1}{2^i}$ of the input resolution. The blocks in the same level have an identical architecture. Inspired by NAS-FPN~\cite{nasfpn}, we define 5 output blocks from $L_3$ to $L_7$ and a $1\times1$ convolution attached to each output block to produce multi-scale features $P_3$ to $P_7$ with the same feature dimension. The rest of the building blocks are used as intermediate blocks before the output blocks. In Neural Architecture Search, we first search for scale permutations for the intermediate and output blocks then determine cross-scale connections between blocks. We further improve the model by adding block adjustments in the search space.

\subsection{Search Space} \label{sec:searchspace}
\paragraph{Scale permutations:} The orderings of blocks are important because a block can only connect to its parent blocks which have lower orderings. We define the search space of scale permutations by permuting intermediate and output blocks respectively, resulting in a search space size of $(N-5)!5!$.
The scale permutations are first determined before searching for the rest of the architecture.
\vspace{-5mm}
\paragraph{Cross-scale connections:} We define two input connections for each block in the search space. The parent blocks can be any block with a lower ordering or block from the stem network. Resampling spatial and feature dimensions is needed when connecting blocks in different feature levels. The search space has a size of $\prod_{i=m}^{N+m-1} C^i_2$, where $m$ is the number of candidate blocks in the stem network.
\vspace{-3mm}
\paragraph{Block adjustments:} We allow block to adjust its scale level and type. The intermediate blocks can adjust levels by $\{-1, 0, 1, 2\}$, resulting in a search space size of $4^{N-5}$. 
All blocks are allowed to select one between the two options \{\textit{bottleneck block}, \textit{residual block}\} described in~\cite{resnet}, resulting in a search space size of $2^{N}$.

\subsection{Resampling in Cross-scale Connections}\label{sec:resampling}
\begin{figure}[t]
\centering
\includegraphics[width=1.0\columnwidth]{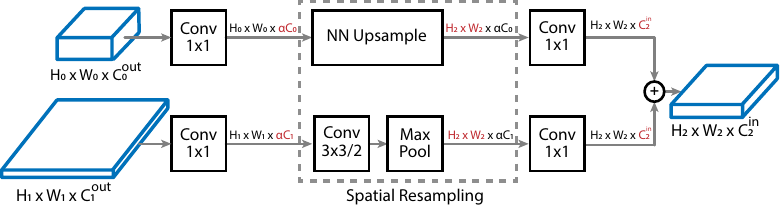}
\caption{\textbf{Resampling operations.} Spatial resampling to upsample (top) and to downsample (bottom) input features followed by resampling in feature dimension before feature fusion.}
\label{fig:resample}
\vspace{-2mm}
\end{figure}

One challenge in cross-scale feature fusion is that the resolution and feature dimension may be different among parent and target blocks. In such case, we perform spatial and feature resampling to match the resolution and feature dimension to the target block, as shown in detail in Figure~\ref{fig:resample}. Here, $C$ is the feature dimension of $3\times3$ convolution in residual or bottleneck block. We use $C^{in}$ and $C^{out}$ to indicate the input and output dimension of a block. For bottleneck block, $C^{in}=C^{out}=4C$; and for residual block, $C^{in}=C^{out}=C$. As it is important to keep the computational cost in resampling low, we introduce a scaling factor $\alpha$ (default value $0.5$) to adjust the output feature dimension $C^{out}$ in a parent block to $\alpha C$. Then, we use a nearest-neighbor interpolation for up-sampling or a stride-2 $3\times3$ convolution (followed by stride-2 max poolings if necessary) for down-sampling feature map to match to the target resolution. Finally, a $1\times1$ convolution is applied to match feature dimension $\alpha C$  to the target feature dimension $C^{in}$. Following FPN~\cite{fpn}, we merge the two resampled input feature maps with elemental-wise addition.

\subsection{Scale-Permuted Model by Permuting ResNet} \label{sec:evolve}
Here we build scale-permuted models by permuting feature blocks in ResNet architecture. The idea is to have a fair comparison between scale-permuted model and scale-decreased model when using the same building blocks. We make small adaptation for scale-permuted models to generate multi-scale outputs by replacing one $L_5$ block in ResNet with one $L_6$ and one $L_7$ blocks and set the feature dimension to $256$ for $L_5$, $L_6$, and $L_7$ blocks. In addition to comparing fully scale-decreased and scale-permuted model, we create a family of models that gradually shifts the model from the scale-decreased stem network to the scale-permuted network. Table~\ref{tab:budget} shows an overview of block allocation of models in the family. We use R[$N$]-SP[$M$] to indicate $N$ feature layers in the handcrafted stem network and $M$ feature layers in the learned scale-permuted network.

For a fair comparison, we constrain the search space to only include scale permutations and cross-scale connections. Then we use reinforcement learning to train a controller to generate model architectures. similar to~\cite{nasfpn}, for intermediate blocks that do not connect to any block with a higher ordering in the generated architecture, we connect them to the output block at the corresponding level. Note that the cross-scale connections only introduce small computation overhead, as discussed in Section~\ref{sec:resampling}. As a result, all models in the family have similar computation as ResNet-50. Figure~\ref{fig:architectures} shows a selection of learned model architectures in the family.

\begin{table}[t]\centering
\small
\setlength\tabcolsep{2.5pt}
\begin{tabular}{l|c|c}
\Xhline{1.0pt}
 & \text{stem network} & \text{scale-permuted network} \\
 & $\{L_2,L_3,L_4,L_5\}$ & $\{L_2,L_3,L_4,L_5,L_6,L_7\}$ \\
 \Xhline{1.0pt}
 R50 & $\{3,4,6,3\}$ & $\{-\}$ \\
  
 R35-SP18 & $\{2,3,5,1\}$ & $\{1,1,1,1,1,1\}$ \\
 
 R23-SP30 & $\{2,2,2,1\}$ & $\{1,2,4,1,1,1\}$ \\
 
 R14-SP39 & $\{1,1,1,1\}$ & $\{2,3,5,1,1,1\}$ \\
 
 R0-SP53 & $\{2,0,0,0\}$ & $\{1,4,6,2,1,1\}$ \\
\Xhline{1.0pt}
 SpineNet-49 & $\{2,0,0,0\}$ & $\{1,2,4,4,2,2\}$ \\
\Xhline{1.0pt}
\end{tabular}
\caption{\textbf{Number of blocks per level for stem and scale-permuted networks.} The scale-permuted network is built on top of a scale-decreased stem network as shown in Figure~\ref{fig:architectures}. The size of scale-decreased stem network is gradually decreased to show the effectiveness of scale-permuted network.}
\label{tab:budget}
\vspace{-2mm}
\end{table}

\subsection{SpineNet Architectures}\label{modela1}
To this end, we design scale-permuted models with a fair comparison to ResNet. However, using ResNet-50 building blocks may not be an optimal choice for building scale-permuted models. We suspect the optimal model may have different feature resolution and block type distributions than ResNet. Therefore, we further include additional block adjustments in the search space as proposed in Section~\ref{sec:searchspace}. The learned model architecture is named SpineNet-49, of which the architecture is shown in Figure~\ref{fig:sub-forth} and the number of blocks per level is given in Table~\ref{tab:budget}.  

\begin{figure}[t]
\vspace*{-0mm}
\centering
\includegraphics[width=0.6\columnwidth]{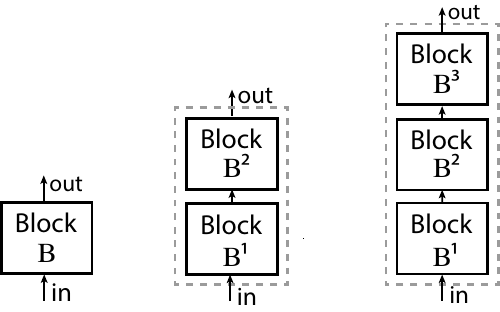}
\caption{\textbf{Increase model depth by block repeat.} From left to right: blocks in SpineNet-49, SpineNet-96, and SpineNet-143.}
\label{fig:repeat}
\vspace{-2mm}
\end{figure}

Based on SpineNet-49, we construct four architectures in the SpineNet family where the models perform well for a wide range of latency-performance trade-offs. The models are denoted as SpineNet-49S/96/143/190: SpineNet-49S has the same architecture as SpineNet-49 but the feature dimensions in the entire network are scaled down uniformly by a factor of $0.65$. SpineNet-96 doubles the model size by repeating each block $\mathbf{B}_k$ twice. The building block $\mathbf{B}_k$ is duplicated into $\mathbf{B}_k^1$ and $\mathbf{B}_k^2$, which are then sequentially connected. The first block $\mathbf{B}_k^1$ connects to input parent blocks and the last block $\mathbf{B}_k^2$ connects to output target blocks. SpineNet-143 and SpineNet-190 repeat each block 3 and 4 times to grow the model depth and adjust $\alpha$ in the resampling operation to $1.0$. SpineNet-190 further scales up feature dimension uniformly by $1.3$. Figure~\ref{fig:repeat} shows an example of increasing model depth by repeating blocks.

Note we do not apply recent work on new building blocks (\eg, ShuffleNetv2 block used in DetNas~\cite{detnas}) or efficient model scaling~\cite{tan2019efficientnet} to SpineNet. These improvements could be orthogonal to this work. 

\begin{table*}[t]\centering
\small
\begin{tabular}{l c| c c | c c c c c c}
\Xhline{1.0pt}
 \multicolumn{1}{l}{\text{backbone model}} & \text{resolution}& \text{\#FLOPs}$\blacktriangle$ & \text{\#Params} & \text{AP} & \text{AP$_{50}$} & \text{AP$_{75}$} & \text{AP$_{S}$} & \text{AP$_{M}$} & \text{AP$_{L}$}\\[2pt]
 \Xhline{1.0pt}
 \textbf{SpineNet-49S} &  640$\times$640& \textbf{33.8B} & \textbf{11.9M} & \textbf{39.5} & \textbf{59.3} & \textbf{43.1} & \textbf{20.9} & \textbf{42.2} & \textbf{54.3} \\
  \hline
 \textbf{SpineNet-49} & 640$\times$640 & \textbf{85.4B}  & \textbf{28.5M} & \textbf{42.8} & \textbf{62.3} & \textbf{46.1} & \textbf{23.7} & \textbf{45.2} & \textbf{57.3} \\
 R50-FPN & 640$\times$640 & 96.8B & 34.0M & 40.4 & 59.9 & 43.6 & 22.7 & 43.5 & 57.0 \\
 R50-NAS-FPN  & 640$\times$640 & 140.0B & 60.3M & 42.4 & 61.8 & 46.1 & 25.1 & 46.7 & 57.8 \\
 \hline
  \textbf{SpineNet-49} & 896$\times$896 & \textbf{167.4B}  & \textbf{28.5M} & \textbf{45.3} & \textbf{65.1} & \textbf{49.1} & \textbf{27.0} & \textbf{47.9} & \textbf{57.7} \\
  \hline
 \textbf{SpineNet-96} & 1024$\times$1024 & \textbf{265.4B} & \textbf{43.0M} & \textbf{47.1} & \textbf{67.1} & \textbf{51.1} & \textbf{29.1} & \textbf{50.2} & \textbf{59.0} \\
 R101-FPN & 1024$\times$1024 & 325.9B & 53.1M & 43.9 & 63.6 & 47.6 & 26.8 & 47.6 & 57.0\\
 \hline
 \textbf{SpineNet-143} & 1280$\times$1280 & \textbf{524.4B} & \textbf{66.9M} & \textbf{48.1} & \textbf{67.6} & \textbf{52.0} & \textbf{30.2} & \textbf{51.1} & \textbf{59.9} \\
 R152-FPN & 1280$\times$1280 & 630.5B & 68.7M & 45.1 & 64.6 & 48.7 & 28.4 & 48.8 & 58.2\\
 \hline
 \hline
 R50-FPN$^\dagger$ & 640$\times$640 & 96.8B & 34.0M & 42.3 & 61.9 & 45.9 & 23.9 & 46.1 & 58.5\\
 \hline
 \hline
 \textbf{SpineNet-49S}$^\dagger$ &  640$\times$640& \textbf{33.8B} & \textbf{12.0M} & \textbf{41.5} & \textbf{60.5} & \textbf{44.6} & \textbf{23.3} & \textbf{45.0} & \textbf{58.0} \\
 \textbf{SpineNet-49}$^\dagger$ & 640$\times$640 & \textbf{85.4B}  & \textbf{28.5M} & \textbf{44.3} & \textbf{63.8} & \textbf{47.6} & \textbf{25.9} & \textbf{47.7} & \textbf{61.1} \\
 \textbf{SpineNet-49}$^\dagger$ & 896$\times$896 & \textbf{167.4B}  & \textbf{28.5M} & \textbf{46.7} & \textbf{66.3} & \textbf{50.6} & \textbf{29.1} & \textbf{50.1} & \textbf{61.7} \\
 \textbf{SpineNet-96}$^\dagger$ & 1024$\times$1024 & \textbf{265.4B} & \textbf{43.0M} & \textbf{48.6} & \textbf{68.4} & \textbf{52.5} & \textbf{32.0} & \textbf{52.3} & \textbf{62.0} \\
 \textbf{SpineNet-143}$^\dagger$ & 1280$\times$1280 & \textbf{524.4B} & \textbf{66.9M} & \textbf{50.7} & \textbf{70.4} & \textbf{54.9} & \textbf{33.6} & \textbf{53.9} & \textbf{62.1} \\
 \textbf{SpineNet-190}$^\dagger$ & 1280$\times$1280 & \textbf{1885.0B} & \textbf{163.6M} & \textbf{52.1} & \textbf{71.8} & \textbf{56.5} & \textbf{35.4} & \textbf{55.0} & \textbf{63.6} \\
\Xhline{1.0pt}
\end{tabular}
\vspace*{-0mm}
\caption{\textbf{One-stage object detection results on COCO \texttt{test-dev}.} We compare employing different backbones with RetinaNet on single model without test-time augmentation. By default we apply protocol B with multi-scale training and ReLU activation to train all models in this table, as described in Section~\ref{sec:settings}. Models marked by dagger ($^\dagger$) are trained with protocol C by applying stochastic depth and swish activation for a longer training schedule. FLOPs is represented by Multi-Adds.}
\label{tab:mainresults}
\vspace{-2mm}
\end{table*}

\begin{table}[t]\centering
\small
\begin{tabular}{l|c|c c}
\Xhline{1.0pt}
 \multicolumn{1}{c|}{\text{model}} & \text{block adju.} &  \text{\#FLOPs} & AP \\
 \Xhline{1.0pt}
 R50-FPN & - & 96.8B & 37.8 \\
 R35-SP18 & - & 91.7B & 38.7  \\
 R23-SP30 & - & 96.5B & 39.7 \\
 R14-SP39 & - & 99.7B & 39.6 \\
 R0-SP53 & - & 95.2B & 40.7  \\
 SpineNet-49 &\cmark & 85.4B & 40.8 \\
\Xhline{1.0pt}
\end{tabular}
\caption{Results comparisons between R50-FPN and scale-permuted models on COCO \texttt{val2017} by adopting protocol A. The performance improves with more computation being allocated to scale-permuted network. We also show the efficiency improvement by having scale and block type adjustments in Section~\ref{sec:searchspace}.}
\label{tab:evolveresults}
\end{table}

\section{Applications}

\subsection{Object Detection}
The SpineNet architecture is learned with RetinaNet detector by simply replacing the default ResNet-FPN backbone model.
To employ SpineNet in RetinaNet, we follow the architecture design for the class and box subnets in~\cite{retinanet}: For SpineNet-49S, we use 4 shared convolutional layers at feature dimension $128$; For SpineNet-49/96/143, we use 4 shared convolutional layers at feature dimension $256$; For SpineNet-190, we scale up subnets by using 7 shared convolutional layers at feature dimension $512$.
To employ SpineNet in Mask R-CNN, we follow the same architecture design in~\cite{rethinking}: For SpineNet-49S/49/96/143, we use 1 shared convolutional layers at feature dimension $256$ for RPN, 4 shared convolutional layers at feature dimension $256$ followed by a fully-connected layers of $1024$ units for detection branch, and 4 shared convolutional layers at feature dimension $256$ for mask branch. For SpineNet-49S, we use $128$ feature dimension for convolutional layers in subnets. For SpineNet-190, we scale up detection subnets by using 7 convolutional layers at feature dimension $384$.



\begin{table}[t]\centering
\small
\begin{tabular}{l | c | c | c}
\Xhline{1.0pt}
 \multicolumn{1}{c|}{\text{model}} & \text{resolution} & \text{AP} & \text{inference latency}\\
 \Xhline{1pt}
 \text{SpineNet-49S} & 640$\times$640  &  39.9  & 11.7ms \\
 \text{SpineNet-49}  &  640$\times$640  & 42.8 & 15.3ms \\
 \text{SpineNet-49} & 896$\times$896 & 45.3 & 34.3ms  \\
 \Xhline{1pt}
\end{tabular}
\caption{Inference latency of RetinaNet with SpineNet on a V100 GPU with NVIDIA TensorRT. Latency is measured for an end-to-end object detection pipeline including pre-processing, detection generation, and post-processing (\eg, NMS).}
\label{tab:realtime}
\vspace{-2mm}
\end{table}

\begin{table*}[t]\centering
\small
\begin{tabular}{l c|c c|c c c c }
\Xhline{1.0pt}
   \multicolumn{1}{l}{backbone model} & resolution & \text{\#FLOPs$\blacktriangle$} & \text{\#Params}
 & \text{AP$_\text{val}$} & \text{AP$^{\text{mask}}_\text{val}$} & \text{AP$_\text{test-dev}$} & \text{AP$^{\text{mask}}_\text{test-dev}$}\\[2pt]
 \Xhline{1.0pt}
  \textbf{SpineNet-49S} & 640$\times$640 & \textbf{60.2B} & \textbf{13.9M} & \textbf{39.3} &\textbf{34.8}&-&-\\
  \hline
  \textbf{SpineNet-49} & 640$\times$640 & \textbf{216.1B} & \textbf{40.8M} & \textbf{42.9} &\textbf{38.1}&-&-\\
  R50-FPN & 640$\times$640 & 227.7B & 46.3M & 42.7 & 37.8 &-&-\\
  \hline
  \textbf{SpineNet-96} & 1024$\times$1024 & \textbf{315.0B} &  \textbf{55.2M} & \textbf{47.2}  & \textbf{41.5} &-&- \\
  R101-FPN & 1024$\times$1024 & 375.5B & 65.3M & 46.6 &41.2&-&-\\
  \hline
  \textbf{SpineNet-143} & 1280$\times$1280 & \textbf{498.8B} & \textbf{79.2M} & \textbf{48.8} &\textbf{42.7}&-&-\\
  R152-FPN & 1280$\times$1280 & 605.3B & 80.9M & 48.1 &42.4&-&-\\
  \hline
  \textbf{SpineNet-190}$^\dagger$ & 1536$\times$1536 & \textbf{2076.8B} & \textbf{176.2M} & \textbf{52.2} & \textbf{46.1} &\textbf{52.5} &\textbf{46.3}\\
\Xhline{1.0pt}

\end{tabular}
\caption{\textbf{Two-stage object detection and instance segmentation results.} 
We compare employing different backbones with Mask R-CNN using $1000$ proposals on single model without test-time augmentation. By default we apply protocol B with multi-scale training and ReLU activation to train all models in this table, as described in Section~\ref{sec:settings}. SpineNet-190 (marked by $^\dagger$) is trained with protocol C by applying stochastic depth and swish activation for a longer training schedule.
FLOPs is represented by Multi-Adds.}
\label{tab:instanceseg}
\vspace*{-2mm}
\end{table*}

\subsection{Image Classification}
To demonstrate SpineNet has the potential to generalize to other visual recognition tasks, we apply SpineNet to image classification.
We utilize the same $P_3$ to $P_7$ feature pyramid to construct the classification network.
Specifically, the final feature map $P = \frac{1}{5} \sum_{i=3}^7 \mathcal{U}(P_i)$ is generated by upsampling and averaging the feature maps, where $\mathcal{U}(\cdot)$ is the nearest-neighbor upsampling to ensure all feature maps have the same scale as the largest feature map $P_3$.
The standard global average pooling on $P$ is applied to produce a $256$-dimensional feature vector followed by a linear classifier with softmax for classification.

\section{Experiments} \label{sec:exp}
For object detection, we evaluate SpineNet on COCO dataset~\cite{coco}. All the models are trained on the \texttt{train2017} split. We report our main results with COCO AP on the \texttt{test-dev} split and others on the \texttt{val2017} split.
For image classification, we train SpineNet on ImageNet ILSVRC-2012~\cite{ilsvrc} and iNaturalist-2017~\cite{inaturalist} and report Top-1 and Top-5 validation accuracy.

\subsection{Experimental Settings}
\label{sec:settings}

\paragraph{Training data pre-processing:} For object detection, we feed a larger image, from 640 to 896, 1024, 1280, to a larger SpineNet. The long side of an image is resized to the target size then the short side is padded with zeros to make a square image.
For image classification, we use the standard input size of $224\times224$.
During training, we adopt standard data augmentation (scale and aspect ratio augmentation, random cropping and horizontal flipping).

\paragraph{Training details:} For object detection, we generally follow~\cite{retinanet,nasfpn} to adopt the same training protocol, denoting as protocol A, to train SpineNet and ResNet-FPN models for controlled experiments described in Figure~\ref{fig:architectures}. In brief, we use stochastic gradient descent to train on Cloud TPU v3 devices with 4e-5 weight decay and 0.9 momentum. All models are \textit{trained from scratch} on COCO \texttt{train2017} with 256 batch size for 250 epochs. The initial learning rate is set to 0.28 and a linear warmup is applied in the first 5 epochs. We apply stepwise learning rate that decays to $0.1\times$ and $0.01\times$ at the last 30 and 10 epoch. We follow~\cite{rethinking} to apply synchronized batch normalization with 0.99 momentum followed by ReLU and implement DropBlock~\cite{dropblock} for regularization.
We apply multi-scale training with a random scale between $[0.8,1.2]$ as in~\cite{nasfpn}. We set base anchor size to 3 for SpineNet-96 or smaller models and 4 for SpineNet-143 or larger models in RetinaNet implementation. 
For our reported results, we adopt an improved training protocol denoting as protocol B. For simplicity, protocol B removes DropBlock and apply stronger multi-scale training with a random scale between $[0.5,2.0]$ for 350 epochs.
To obtain the most competitive results, we add stochastic depth with keep prob 0.8~\cite{dropconnect} for stronger regularization and replace ReLU with swish activation~\cite{swish} to train all models for 500 epochs, denoting as protocol C. We also adopt a more aggressive multi-scale training strategy with a random scale between [0.1, 2.0] for SpineNet-143/190 when using protocol C.
For image classification, all models are trained with a batch size of $4096$ for $200$ epochs. We used cosine learning rate decay~\cite{he2019bag} with linear scaling of learning rate and gradual warmup in the first $5$ epochs~\cite{goyal2017accurate}.

\paragraph{NAS details:} We implement the recurrent neural network based controller proposed in~\cite{nas} for architecture search, as it is the only method we are aware of that supports searching for permutations. We reserve 7392 images from \texttt{train2017} as the validation set for searching. To speed up the searching process, we design a proxy SpineNet by uniformly scaling down the feature dimension of SpineNet-49 with a factor of 0.25, setting $\alpha$ in resampling to 0.25, and using feature dimension 64 in the box and class nets. To prevent the search space from growing exponentially, we restrict intermediate blocks to search for parent blocks within the last 5 blocks built and allow output blocks to search from all existing blocks. At each sample, a proxy task is trained at image resolution 512 for 5 epochs. AP of the proxy task on the reserved validation set is collected as reward. The controller uses $100$ Cloud TPU v3 in parallel to sample child models. The best architectures for R35-SP18, R23-SP30, R14-SP39, R0-SP53, and SpineNet-49 are found after 6k, 10k, 13k, 13k, and 14k architectures are sampled. 

\subsection{Learned Scale-Permuted Architectures}
In Figure~\ref{fig:architectures}, we observe scale-permuted models have permutations such that the intermediate features undergo the transformations that constantly up-sample and down-sample feature maps, showing a big difference compared to a scale-decreased backbone. It is very common that two adjacent intermediate blocks are connected to form a \textit{deep} pathway. The output blocks demonstrate a different behavior preferring longer range connections. In Section~\ref{sec:ablation}, we conduct ablation study to show the importance of learned scale permutation and connections.

\subsection{ResNet-FPN vs. SpineNet}
We first present the object detection results of the 4 scale-permuted models discussed in Section~\ref{sec:evolve} and compare with the ResNet50-FPN baseline. The results in Table~\ref{tab:evolveresults} support our claims that: (1) The scale-decreased backbone model is not a good design of backbone model for object detection; (2) allocating computation on the proposed scale-permuted model yields higher performance.

Compared to the R50-FPN baseline, R0-SP53 uses similar building blocks and gains $2.9\%$ AP with a learned scale permutations and cross-scale connections. The SpineNet-49 model further improves efficiency by reducing FLOPs by 10\% while achieving the same accuracy as R0-SP53 by adding scale and block type adjustments.

\begin{table*}[t]
\small
\begin{center}
\begin{tabular}{ l|rrcc|rrcc }
\Xhline{1.0pt}
 \multicolumn{1}{c|}{\multirow{2}{*}{network}} & \multicolumn{4}{c|}{ImageNet ILSVRC-2012 (1000-class)} & \multicolumn{4}{c}{iNaturalist-2017 (5089-class)} \\
\cline{2-9}
 & \multicolumn{1}{c}{\text{\#FLOPs}$\blacktriangle$} & \multicolumn{1}{c}{\text{\#Params}} & \multicolumn{1}{c}{Top-1 \%} & \multicolumn{1}{c|}{Top-5 \%} & \multicolumn{1}{c}{\text{\#FLOPs}} & \multicolumn{1}{c}{\text{\#Params}} & \multicolumn{1}{c}{Top-1 \%} & \multicolumn{1}{c}{Top-5 \%} \\
\Xhline{1.0pt}
 \textbf{SpineNet-49} & \textbf{3.5B} & \textbf{22.1M} & \textbf{77.0} & \multicolumn{1}{c|}{\textbf{93.3}} & \textbf{3.5B} & \textbf{23.1M} & \textbf{59.3} & \textbf{81.9}\\
 ResNet-34 & 3.7B & 21.8M & 74.4 & \multicolumn{1}{c|}{92.0} & 3.7B & 23.9M & 54.1 & 76.7\\
 ResNet-50 & 4.1B & 25.6M & 77.1 & \multicolumn{1}{c|}{93.6} & 4.1B & 33.9M & 54.6 & 77.2\\
 \hline
 \textbf{SpineNet-96} & \textbf{5.7B} & \textbf{36.5M} & \textbf{78.2} & \multicolumn{1}{c|}{\textbf{94.0}} & \textbf{5.7B} & \textbf{37.6M} & \textbf{61.7} & \textbf{83.4}\\
 ResNet-101 & 7.8B & 44.6M & 78.2 & \multicolumn{1}{c|}{94.2} & 7.8B & 52.9M & 57.0 & 79.3\\
 \hline
 \textbf{SpineNet-143} & \textbf{9.1B} & \textbf{60.5M} & \textbf{79.0} & \multicolumn{1}{c|}{\textbf{94.4}} & \textbf{9.1B} & \textbf{61.6M} & \textbf{63.6} & \textbf{84.8}\\
 ResNet-152 & 11.5B & 60.2M & 78.7 & \multicolumn{1}{c|}{94.2} & 11.5B & 68.6M & 58.4 & 80.2\\
\Xhline{1.0pt}
\end{tabular}
\end{center}
\vspace*{-2mm}
\caption{\textbf{Image classification results on ImageNet and iNaturalist.} Networks are sorted by increasing number of FLOPs. Note that the penultimate layer in ResNet outputs a 2048-dimensional feature vector for the classifier while SpineNet's feature vector only has 256 dimensions. Therefore, on iNaturalist, ResNet and SpineNet have around 8M and 1M more parameters respectively.}
\label{tab:classification}
\vspace*{-2mm}
\end{table*}

\subsection{Object Detection Results}
\paragraph{RetinaNet:}
We evaluate SpineNet architectures on the COCO bounding box detection task with a RetinaNet detector. The results are summarized in Table~\ref{tab:mainresults}. SpineNet models outperform other popular detectors by large margins, such as ResNet-FPN, and NAS-FPN at various model sizes in both accuracy and efficiency. Our largest SpineNet-190 achieves 52.1\% AP on single model object detection without test-time augmentation.
\vspace{-3mm}

\paragraph{Mask R-CNN:} \label{exp_mrcnn}
We also show results of Mask R-CNN models with different backbones for COCO instance segmentation task.
Being consistent with RetinaNet results,
SpineNet based models are able to achieve better AP and mask AP with smaller model size and less number of FLOPs. Note that SpineNet is learned on box detection with RetinaNet but works well with Mask R-CNN.
\vspace{-5mm}

\paragraph{Real-time Object Detection:}
Our SpineNet-49S and SpineNet-49 with RetinaNet run at 30+ fps with NVIDIA TensorRT on a V100 GPU. We measure inference latency using an end-to-end object detection pipeline including pre-processing, bounding box and class score generation, and post-processing with non-maximum suppression, reported in Table~\ref{tab:realtime}.

\begin{table}[t]\centering
\small
\begin{tabular}{c| c |c}
\Xhline{1.0pt}
 \text{model shape} & \text{fixed block ordering}
 & AP\\
 \Xhline{1pt}
 \text{Hourglass}  & 
 \begin{tabular}{@{}c@{}} \{$3L_2, 3L_3, 5L_4, 1L_5, 1L_7, 1L_6,$\\  $1L_5, 1L_4, 1L_3$\}\end{tabular}
 & 38.3\% \\
 \hline
 \text{Fish}  &
 \begin{tabular}{@{}c@{}} \{$2L_2, 2L_3, 3L_4, 1L_5, 2L_4, 1L_3,$ \\$1L_2, 1L_3, 1L_4, 1L_5, 1L_6,1L_7$\}\end{tabular}
  & 37.5\% \\
 \Xhline{1pt}
 \text{R0-SP53}   & - & 40.7\% \\
 \Xhline{1pt}
\end{tabular}
\caption{\textbf{Importance of learned scale permutation.} We compare our R0-SP53 model to hourglass and fish models with fixed block orderings. All models learn the cross-scale connections by NAS.}
\label{tab:fixedorder}
\end{table}

\begin{table}[t]\centering
\small
\begin{tabular}{c| c c c|c}
\Xhline{1.0pt}
 \text{model} & long & short & sequential & AP\\
 \Xhline{1pt}
 \text{R0-SP53}  & \cmark & \cmark & - & 40.7\% \\
 \text{Graph damage (1)}  & \cmark & \xmark & - & 35.8\% \\
 \text{Graph damage (2)}  & \xmark & \cmark & - & 28.6\% \\
 \text{Graph damage (3)}  & \xmark & \xmark & \cmark & 28.2\% \\
 \Xhline{1pt}
\end{tabular}
\caption{\textbf{Importance of learned cross-scale connections.} We quantify the importance of learned cross-scale connections by performing three graph damages by removing edges of: (1) short-range connections; (2) long-range connections; (3) all connections then sequentially connecting every pair of adjacent blocks.}
\label{tab:damage}
\vspace{-2mm}
\end{table}

\subsection{Ablation Studies}
\label{sec:ablation}

\paragraph{Importance of Scale Permutation:}
We study the importance of learning scale permutations by comparing learned scale permutations to fixed ordering feature scales. We choose two popular architecture shapes in encoder-decoder networks: (1) A \textit{Hourglass} shape inspired by~\cite{stackedhourglass, fpn}; (2) A \textit{Fish} shape inspired by~\cite{fishnet}. Table~\ref{tab:fixedorder} shows the ordering of feature blocks in the \textit{Hourglass} shape and the \textit{Fish} shape architectures. Then, we learn cross-scale connections using the same search space described in Section~\ref{sec:searchspace}. The performance shows jointly learning scale permutations and cross-scale connections is better than only learning connections with a fixed architecture shape. Note there may exist some architecture variants to make \textit{Hourglass} and \textit{Fish} shape model perform better, but we only experiment with one of the simplest fixed scale orderings.
\vspace{-2mm}

\paragraph{Importance of Cross-scale Connections:}
The cross-scale connections play a crucial role in fusing features at different resolutions throughout a scale-permuted network. We study its importance by graph damage. For each block in the scale-permuted network of R0-SP53, cross-scale connections are damaged in three ways: (1) Removing the short-range connection; (2) Removing the long-range connection; (3) Removing both connections then connecting one block to its previous block via a sequential connection. In all three cases, one block only connects to one other block. In Table~\ref{tab:damage}, we show scale-permuted network is sensitive to any of edge removal techniques proposed here. The (2) and (3) yield severer damage than (1), which is possibly because of short-range connection or sequential connection can not effectively handle the frequent resolution changes. 

\begin{table*}[t]\centering
\small
\setlength\tabcolsep{5.5pt}
\begin{tabular}{l| r r | c c c c}
\Xhline{1.0pt}
 \multicolumn{1}{c|}{\text{backbone model}} & \text{\#FLOPs} & \text{\#Params} & \text{AP} & \text{AP$_{S}$} & \text{AP$_{M}$} & \text{AP$_{L}$}\\[2pt]
 \Xhline{1.0pt}
 \bf SpineNet-49XS (MBConv) & \textbf{0.17B} & \textbf{0.82M} & \textbf{17.5} & \textbf{2.3} & \textbf{17.2} & \textbf{33.6} \\
  MobileNetV3-Small-SSDLite~\cite{mobilenetv3}  & 0.16B & 1.77M & 16.1 & - & - & - \\
  \hline
  \bf SpineNet-49S (MBConv) & \textbf{0.52B} & \textbf{0.97M} & \textbf{24.3} & \textbf{7.2} & \textbf{26.2} & \textbf{41.1} \\
  MobileNetV3-SSDLite~\cite{mobilenetv3}  & 0.51B & 3.22M & 22.0 & - & - & - \\
  MobileNetV2-SSDLite~\cite{mobilenetv2}  & 0.80B & 4.30M & 22.1 & - & - & - \\
  MnasNet-A1-SSDLite~\cite{mnasnet}  & 0.80B & 4.90M & 23.0 & 3.8 & 21.7 & 42.0 \\
 \hline
 \bf SpineNet-49 (MBConv) & \textbf{1.00B} & \textbf{2.32M} & \textbf{28.6} & \textbf{9.2} & \textbf{31.5} & \textbf{47.0}  \\
 MobileNetV2-NAS-FPNLite (7 @64)~\cite{nasfpn}  & 0.98B & 2.62M & 25.7 & - & - & - \\
 MobileNetV2-FPNLite~\cite{mobilenetv2}  & 1.01B & 2.20M & 24.3 & - & - & - \\
\Xhline{1.0pt}
\end{tabular}
\caption{\textbf{Mobile-size object detection results.} We report single model results without test-time augmentation on COCO \texttt{test-dev}.}
\label{tab:ondeviceresults}
\end{table*}


\begin{table*}[t]
\small
\begin{center}
\begin{tabular}{ l|rrcc|rrcc }
\Xhline{1.0pt}
 \multicolumn{1}{c|}{\multirow{2}{*}{network}} & \multicolumn{4}{c|}{ImageNet ILSVRC-2012 (1000-class)} & \multicolumn{4}{c}{iNaturalist-2017 (5089-class)} \\
\cline{2-9}
 & \multicolumn{1}{c}{\text{\#FLOPs}$\blacktriangle$} & \multicolumn{1}{c}{\text{\#Params}} & \multicolumn{1}{c}{Top-1 \%} & \multicolumn{1}{c|}{Top-5 \%} & \multicolumn{1}{c}{\text{\#FLOPs}} & \multicolumn{1}{c}{\text{\#Params}} & \multicolumn{1}{c}{Top-1 \%} & \multicolumn{1}{c}{Top-5 \%} \\
\Xhline{1.0pt}
 SpineNet-49 & \multicolumn{1}{r}{\multirow{2}{*}{3.5B}} & \multicolumn{1}{r}{\multirow{2}{*}{22.1M}} & 77.0 & \multicolumn{1}{c|}{93.3} & \multicolumn{1}{r}{\multirow{2}{*}{3.5B}} & \multicolumn{1}{r}{\multirow{2}{*}{23.1M}} & 59.3 & 81.9\\
 SpineNet-49$^\dagger$ & & & 78.1 & 94.0 & & & 63.3 & 85.1\\
 \hline
 SpineNet-96 &  \multicolumn{1}{r}{\multirow{2}{*}{5.7B}} & \multicolumn{1}{r}{\multirow{2}{*}{36.5M}} & 78.2 & \multicolumn{1}{c|}{94.0} & \multicolumn{1}{r}{\multirow{2}{*}{5.7B}} & \multicolumn{1}{r}{\multirow{2}{*}{37.6M}} & 61.7 & 83.4\\
 SpineNet-96$^\dagger$ & & & 79.4 & 94.6 & & & 64.7 & 85.9\\
 \hline
 SpineNet-143 & \multicolumn{1}{r}{\multirow{2}{*}{9.1B}} & \multicolumn{1}{r}{\multirow{2}{*}{60.5M}} & 79.0 & \multicolumn{1}{c|}{94.4} & \multicolumn{1}{r}{\multirow{2}{*}{9.1B}} & \multicolumn{1}{r}{\multirow{2}{*}{61.6M}} & 63.6 & 84.8\\
 SpineNet-143$^\dagger$  & & & 80.1 & 95.0 & & & 66.7 & 87.1\\
 \hline
 SpineNet-190$^\dagger$ & 19.1B & 127.1M & 80.8 & \multicolumn{1}{c|}{95.3} & 19.1B & 129.2M & 67.6 & 87.4\\
\Xhline{1.0pt}
\end{tabular}
\end{center}
\caption{The performance of SpineNet classification model can be further improved with a better training protocol by 1) adding stochastic depth, 2) replacing ReLU with swish activation and 3) using label smoothing of 0.1 (marked by $^\dagger$).}
\label{tab:classification_best}
\end{table*}

\subsection{Image Classification with SpineNet}
Table~\ref{tab:classification} shows the image classification results.
Under the same setting, SpineNet's performance is on par with ResNet on ImageNet but using much fewer FLOPs.
On iNaturalist, SpineNet outperforms ResNet by a large margin of around 5\%.
Note that iNaturalist-2017 is a challenging fine-grained classification dataset containing 579,184 training and 95,986 validation images from 5,089 classes.

To better understand the improvement on iNaturalist, we created iNaturalist-bbox with objects cropped by ground truth bounding boxes collected in~\cite{inaturalist}.
The idea is to create a version of iNaturalist with an iconic single-scaled object centered at each image to better understand the performance improvement.
Specifically, we cropped all available bounding boxes (we enlarge the cropping region to be 1.5$\times$ of the original bounding box width and height to include context around the object), resulted in 496,164 training and 48,736 validation images from 2,854 classes.
On iNaturalist-bbox, the Top-1/Top-5 accuracy is 63.9\%/86.9\% for SpineNet-49 and 59.6\%/83.3\% for ResNet-50, with a 4.3\% improvement in Top-1 accuracy.
The improvement of SpineNet-49 over ResNet-50 in Top-1 is 4.7\% on the original iNaturalist dataset.
Based on the experiment, we believe the improvement on iNaturalist is not due to capturing objects of variant scales but the following 2 reasons: 1) capturing subtle local differences thanks to the multi-scale features in SpineNet; 2) more compact feature representation (256-dimension) that is less likely to overfit.

\section{Conclusion}
In this work, we identify that the conventional scale-decreased model, even with decoder network, is not effective for simultaneous recognition and localization. We propose the scale-permuted model, a new meta-architecture, to address the issue. To prove the effectiveness of scale-permuted models, we learn SpineNet by Neural Architecture Search in object detection and demonstrate it can be used directly in image classification. SpineNet significantly outperforms prior art of detectors by achieving 52.1\% AP on COCO \texttt{test-dev}. The same SpineNet architecture achieves a comparable top-1 accuracy on ImageNet with much fewer FLOPs and 5\% top-1 accuracy improvement on challenging iNaturalist dataset. In the future, we hope the scale-permuted model will become the meta-architecture design of backbones across many visual tasks beyond detection and classification.

\vspace{3mm}
\par\noindent\textbf{Acknowledgments:} We would like to acknowledge Yeqing Li, Youlong Cheng, Jing Li, Jianwei Xie, Russell Power, Hongkun Yu, Chad Richards, Liang-Chieh Chen, Anelia Angelova, and the Google Brain team for their help.


\appendix
\section*{Appendix A: Mobile-size Object Detection}\label{ap:mobile}
For mobile-size object detection, we explore building SpineNet with MBConv blocks using the parametrization proposed in~\cite{mnasnet}, which is the inverted bottleneck block~\cite{mobilenetv2} with SE module~\cite{senet}. Following~\cite{mnasnet}, we set feature dimension $\{16, 24, 40, 80, 112, 112, 112\}$, expansion ratio $6$, and kernel size $3\times3$ for $L_1$ to $L_7$ MBConv blocks. Each block in SpineNet-49 is replaced with the MBConv block at the corresponding level. Similar to~\cite{mnasnet}, we replace the first convolution and maxpooling in stem with a $3\times3$ convolution at feature dimension $8$ and a $L_1$ MBConv block respectively and set the first $L_2$ block to stride $2$. The first $1\times1$ convolution in resampling to adjust feature dimension is removed. All convolutional layers in resampling operations and box/class nets are replaced with separable convolution in order to have comparable computation with MBConv blocks. Feature dimension is reduced to $48$ in the box/class nets. We further construct SpineNet-49XS and SpineNet-49S by scaling the feature dimension of SpineNet-49 by $0.6\times$ and $0.65\times$ and setting the feature dimension in the box/class nets to $24$ and $40$ respectively. We adopt training protocol B with swish activation to train all models with RetinaNet for 600 epochs at resolution $256$ for SpineNet-49XS and $384$ for other models. The results are presented in Table~\ref{tab:ondeviceresults} and the FLOPs \vs AP curve is plotted in Figure~\ref{fig:mobile_detection_performance}. Bulit with MBConv blocks, SpineNet-49XS/49S/49 use less computation but outperform MnasNet, MobileNetV2, and MobileNetV3 by 2-4\% AP.

Note that as all the models in this section use handcrafted MBConv blocks, the performance should be no better than a joint search of SpineNet and MBConv blocks with NAS.


\section*{Appendix B: Image Classification}
Inspired by protocol C, we conduct SpineNet classification experiments using an improved training protocol by 1) adding stochastic depth, 2) replacing ReLU with swish activation and 3) using label smoothing of 0.1.
From results in Table~\ref{tab:classification_best}, we can see that the improved training protocol yields around 1\% Top-1 gain on ImageNet and 3-4\% Top-1 gain on iNaturalist-2017.

{\small
\bibliographystyle{ieee_fullname}
\bibliography{arxiv}
}

\end{document}